\documentclass[10pt,final,conference,compsoc,oneside,letterpaper,twocolumn]{IEEEtran}
\ifCLASSOPTIONcompsoc
  \usepackage[nocompress]{cite}
\else
  \usepackage{cite}
\fi
\ifCLASSINFOpdf
   \usepackage[pdftex]{graphicx}
   \graphicspath{{../pdf/}{../jpeg/}}
   \DeclareGraphicsExtensions{.pdf,.jpeg,.png}
\else
   \usepackage[dvips]{graphicx}
   \graphicspath{{../eps/}}
  \DeclareGraphicsExtensions{.eps}
\fi
\begin{document}
\title{Hyperspectral Imaging Technology and Transfer Learning\\Utilized in Identification Haploid Maize Seeds}

\author{\IEEEauthorblockN{Wen-Xuan Liao\IEEEauthorrefmark{1}, Xuan-Yu Wang\IEEEauthorrefmark{1}, Dong An\IEEEauthorrefmark{1}, Yao-Guang Wei\IEEEauthorrefmark{1}}
\IEEEauthorblockA{\IEEEauthorrefmark{1}College of Information and Electriacl Engineering China Agricultural University,\\Beijing 100083,China \{vane,wangxuanyu,andong,weiyaoguang\}@cau.edu.cn}}

% compsoc conferences compared to traditional conferences), use this
%Telephone: (800) 555--1212, Fax: (888) 555--1212}


\maketitle


\begin{abstract}
It is extremely important to correctly identify the cultivars of maize seeds in the breeding process of maize. In this paper, the transfer learning as a method of deep learning is adopted to establish a model by combining with the hyperspectral imaging technology. The haploid seeds can be recognized from large amount of diploid maize ones with great accuracy through the model. First, the information of maize seeds on each wave band is collected using the hyperspectral imaging technology, and then the recognition model is built on VGG-19 network, which is pre-trained by large-scale computer vision database (Image-Net). The correct identification rate of model utilizing seed spectral images containing 256 wave bands (862.5-1704.2nm) reaches 96.32\%, and the correct identification rate of the model utilizing the seed spectral images containing single-band reaches 95.75\%. The experimental results show that, CNN model which is pre-trained by visible light image database can be applied to the near-infrared hyperspectral imaging-based identification of maize seeds, and high accurate identification rate can be achieved. Meanwhile, when there is small amount of data samples, it can still realize high recognition by using transfer learning. The model not only meets the requirements of breeding recognition, but also greatly reduce the cost occurred in sample collection.
\end{abstract}


% \fi
\IEEEpeerreviewmaketitle

\section{Introduction}
Seed is the foundation of agricultural engineering. Continuously growing population and constantly changing global climate determine that there is a need for more effective breeding strategies for new species, especially to maximumly improve speed and accuracy of the identification and sorting of target seeds in the breeding process. Mixture of different types of seeds will directly result in the reduction of purity in breeding experiment, which further leads to a reduction in the yield of crops. Maize is one of the most widely planted commercial crops in China, the maize haploid breeding technique based on biological induction has become the key to breed new species of maize. Relevant haploid breeding technique helps effectively reduce the breeding time [1], save costs of labor and materials and it is of great importance to germplasm improvement. However, the occurrence rate of maize haploid under natural conditions is about 1\%, while the occurrence rate rises to 8\%-15\% after artificial induction([2],[3],[4]), that is, under the breeding environments, there are just 8\%-15\% of seeds belonging to haploid, while the rest of them are of diploid. Therefore, it has always been the key point to realize rapid, accurate and non-destructive identification and sorting haploid maize seeds.

The widely adopted seed identification methods at present include: morphological method [5], protein electrophoresis [11], DNA molecular marker technology tests [12], genetic marker method [13]. However the first three methods are expensive and time-consuming and requires practiced operators [14]. The genetic marker method is not a good choice for automated machine vision sorting [15]. To collect information without damaging the seeds, the possible technological means is mainly lossless optics, such as machine vision technology [16]. These technologies display limited ability to collect information. Because the information that the machine vision system mainly collects mainly includes the information of external features of seeds, such as color, texture, contour, etc., it seems less effective compared with the near infrared spectroscopy technology with which the spectral features relevant to the chemical contents that the seeds contain are collected [17].

Hyperspectral imaging technology refers that the near infrared imaging is applied at every wavelength point by combining the information of near infrared spectroscopy with the information of near-infrared images, that is, it can be used to generalize the space information and the near infrared spectroscopy information of collected samples. The near infrared spectroscopy is very sensitive to hydrocarbons, hydroxyls, and amines among the organic matters, and correspondingly, information such as proteins, starch, moisture, and fat etc. that the sample contains can be reflected [24] [6]. In recent years, high-spectrum technology makes success in the field of food quality inspection [7], and it is also adopted in the task of seed detection and sorting. [8][9]. Based on above, the hyperspectral imaging technology is feasible to collect the infoemation on the diversity of haploid and diploid maize seeds organics or spatial shapes.

CNN model displays excellent performance in computer vision, and it has been successfully applied in the field of image recognition such as face recognition etc. However, the large-capacity model whose training starts from randomly initialized weights requires a large number of data sets, while it is impossible to provide such a large number of data sets in the maize seeds identification study. These problems can be solved by transfer learning in the field of computer vision and machine learning. Compared with the models that use nothing but small data set with randomly initialized weights, it is proved that this new adjustment process can improve the generalization of the model [10], and the convergence starts when there is just a few of iteration times. Currently, a lot of researchers transfer CNN pre-trained in Image-Net data set to other vision identification tasks. For example, Matthew et al. [21] applied it to Caltech-101 and Caltech-256 image classifications, and Maxime et al.[22] applied it to  Pascal VOC 2007 and 2012 data set image classifications, and Gustavo et al. [23] applied it to medical X-ray image classification. Therefore, CNN model pre-trained through Image-Net not only can be transferred to other visible light data set, but also to the data set of images of other wave bands. At present, there is no research touching upon the possibility of transferring CNN model pre-trained through Image-Net to the data set of hyperspectral images. However, based on investigation, we believe it is feasible. In this paper, the network is initialized using the pre-trained deep learning model available from public sources, and re-training is conducted using relatively smaller training set so that the classification can be realized based on the hyperspectral image of maize seeds.

In this article, the main contributions are as follows:

1.In our study, we design the haploid maize seeds identification model consisting transfer learning technology and hyperspectral imaging technology that samples maize seeds¡¯ embryo surface as images data.

2.The correct recognition rate of maize haploid seeds reaches 96.32\% if the model built on the high-spectrum images of maize seeds containing 256 wave bands is adopted. The average correct recognition rate of maize seeds is 94.08\% and its highest rate reaches 95.85\% if the model built on single-band and high-spectrum images of maize seeds is adopted. Among them, the optimal wave band is identical with the correct recognition rate of full-wave band hyperspectral image. Sampling and modeling with single-band samples data can effectively reduce the cost of breeding sorting, despite expensive full-band samples. Hence, the acquisition cost can be reduced and the speed would be improved through utilizing optical filter covering the full-wave band near-infrared camera.

3.Our study can expend not only to identification of other agricultural products seeds, but also to other fields.
\section{Materials and Method}
\subsection{Dataset}
Samples adopted in the experiment are species called ¡°Zhengdan 958 Maize¡± bred through induction of high-oil crossbreed carrying R1-nj genetic marker, which is developed by Maize Improvement Center of China Agricultural University. Among which, there are 100 haploid seeds and 100 diploid seeds. After each maize seed is dried, dehydrated and numbered, the samples are stored at 5$^{\circ}$C, and their images are obtained using near-infrared hyperspectral imging technology. Considering great difference between embryo surface and non-embryo surface, the image of embryo surface of each seed is adopted in the experiment, that is, the embryo¡¯s surface of each seed is positioned towards the light during acquisition. To maximally reduce the influence of instrumental parameter drifts on the measurement results, the alternate sampling is performed between haploid seed and decreases during the actual sampling process.In the experiment, we will process the collected data as follows: (a) different column matrix is set for labeling the hyperspectral image of 100 haploid seed of Zhengdan 958 and the hyperspectral image of 100 Diploid seed; (b) seeds are randomly selected from haploid and diploid, and the images of these 80 seeds are taken as the test set, while the remaining images are put into the network training as a training set.

\begin{figure}[!t]
\centering
\includegraphics[width=2.5in]{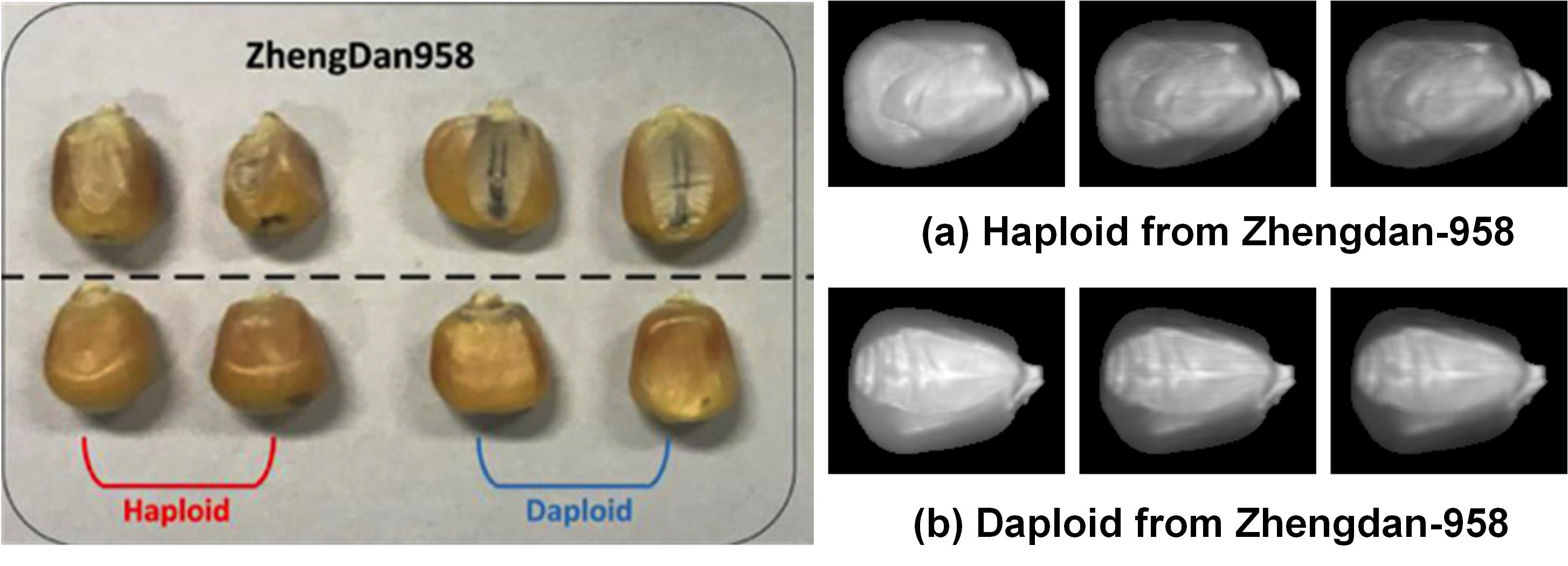}
\caption{Samples of haploid and diploid maize seeds (left) and their hyperspectral images taken at 962.7, 1132.3, 1364.4nm wavebands (right) .}
\label{fig_sim}
\end{figure}
\begin{table}[!t]
	\renewcommand{\arraystretch}{1.3}
	\caption{THE NUMBER, VARIETTY, AND MEASURMENT OF SAMPLE. }
	\label{table_example}
	\centering
	\begin{tabular}{c c c c}
		\hline
		\bfseries Acquisition mode & \bfseries Variety & \bfseries Haploid & \bfseries Diploid \\
		\hline
		HIS & Zhengdan-958 & 100 & 100\\
		\hline
	\end{tabular}
\end{table}

\subsection{Hyperspectral Imaging Technology}
The hyperspectral images were collected by the push-broom GaiaSorter hyperspectral system. The hyperspectral imaging system is mainly comprised of four components, that is, uniform light source, spectral camera, mobile control platform and computer. The uniform light source is comprised of two sets of bromine tungsten lamp, while the light source gives out uniform light through thermal radiation. As for the spectral camera, Image-¦Ë-N17E spectrum near-infrared-enhanced hyperspectral camera (Zolix Instruments Co., Ltd.) is adopted, which integrates the imaging spectrometer of Imspector series and the InGaAs CCD camera, and camera¡¯s spectral region ranges from 862.9 to 1704.2nm £¨containing 256 wave bands£©, which covers near-infrared wave band; its spectral resolution is 5nm, and pixel 320x256, and slit width 30¦Ìm. The system mobile control platform is controlled by stepping motor, and the image acquisition is performed using Spectra View, an image acquisition software. If it is ensured that the image is not distorted, the platform movement speed is set to 0.27 cm/s, and the exposure time is 35 ms. Each image acquired is a three-dimensional image (x, y, $\lambda$), and the collected image was a (320x2000x256) image cube. In order to reduce interference from the external environment, the images are captured inside the camera bellows. As for the measurement errors of hyperspectral image caused by light source fluctuation and dark current is corrected by formula£¨1£©based on black and white reference.\\
\begin{equation}
\label{eqn_example}
R_{cur}=\frac{{R_{sam}}-{R_{dar}}}{{R_{whi}}-{R_{dar}}}
\end{equation}

$R_{cur}$ Denotes calibrated sample images, and $R_{sam}$ denotes original sample images, and $R_{dar}$ denotes dark reference image, and $R_{whi}$ denotes white reference image. The dark reference image can be captured by covering the camera view with lens cap; the white reference image can be captured by replacing measurement object with the pure white board, and completely covering one frame of camera after being lighted. All the calibrated sample images will be applied to the experimental analysis thereafter.
\subsection{Image Segmentation and Characteristics Extraction}
Calibrated sample images still contain the background information during acquisition. To separate the true information of the seed from the background, we have adopted self-adaptive threshold segmentation and masking (Huang et al. 2016) to extract the region of interests (ROIs), and the procedures are follows: (a) within the region of interests (ROIs) extracted on 60 (1064.8nm), the contrast ratio of seeds to the background is the highest. The maximum value of background is selected as the threshold value for image binarization; (b) the peripheric coordinate of each sample is obtained to form a binary image, and the rectangular area of each sample is determined. Binary masks are generated through rectangular regions to obtain ROIs of 256 bands; (c) the true information of the seed is obtained by multiplying each seed¡¯s ROI with its corresponding binary image, so that the interference of background information can be removed. The input layer of VGG-19 network requires a resolution ratio of the image of 224x224, while the resolution ratio of the pre-treated image of seed is 141x111, which does not meet the network requirements. We adopted the method of background filling to modify the seed image into 224x224 (Figure 3), so it satisfies the requirements of the input layer of the network, while it does not change the information contained in the image.\\
The software used to process the seed sample images is MATLAB 2016a (USA, MathWorks Company).
\subsection{Convolutional Neural Network}
In the experiment, convolutional neural networks VGG-19[18] is utilized to extract the features of the seed image and identify haploid. The VGG network was proposed by the Oxford Vision Group and it won the championship in the 2014 ILSVR competition.

VGG-19 is comprised of 19 layers of network, containing 16 convolutional layer and 3 full connection layers. In each convolutional layer, ReLU is taken as activation function. In full connection layer 1 and full connection layer 2, dropout is adopted to prevent over-fitting; finally, softmax is adopted as loss function. The first 18 layers are used for feature extraction and the last layer is used for classification. The network uses nothing but small convolution kernel at 3x3, which is the smallest size capable of capturing all directions and central concept. Multiple convolution layers at 3x3 display greater nonlinearity than one convolutional layer of large size, making the function more deterministic and greatly reducing the number of parameters. Due to network depth, and because small size of convolution kernel helps realize implicit regularization, VGG network starts to converge when there is just a few of iteration times. Therefore, we have selected VGG-19 network in accomplishing the identification task of maize seeds haploid.

\begin{figure*}[!t]
	\centering
	\includegraphics[width=4.0in]{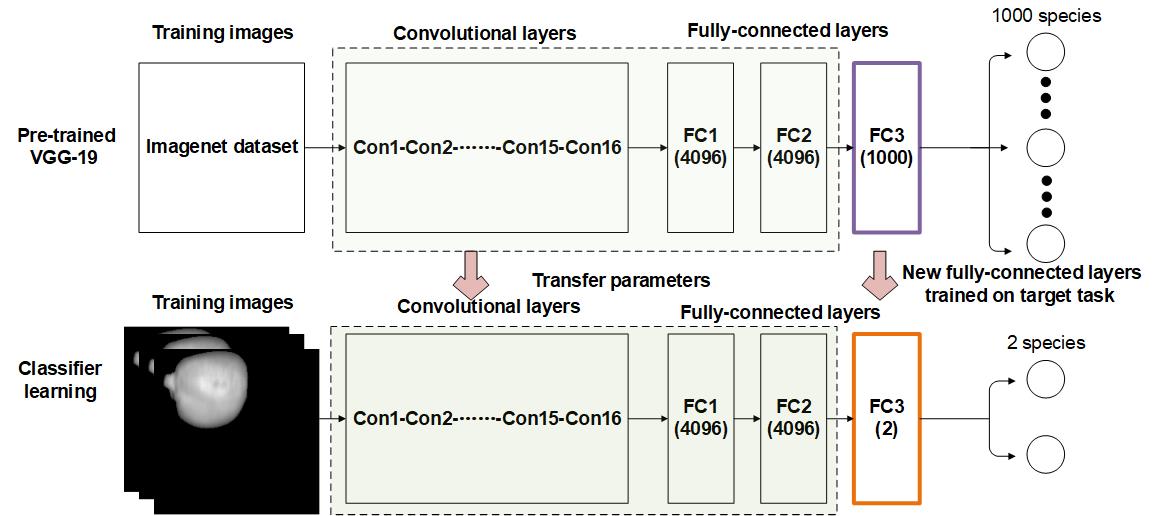}
	\caption{The structure of the modified transfer-learning neutral network improved from Image-Net to classifying maize seeds.}
	\label{fig_sim}
\end{figure*}
\begin{figure*}[!t]
	\centering
	\includegraphics[width=4.0in]{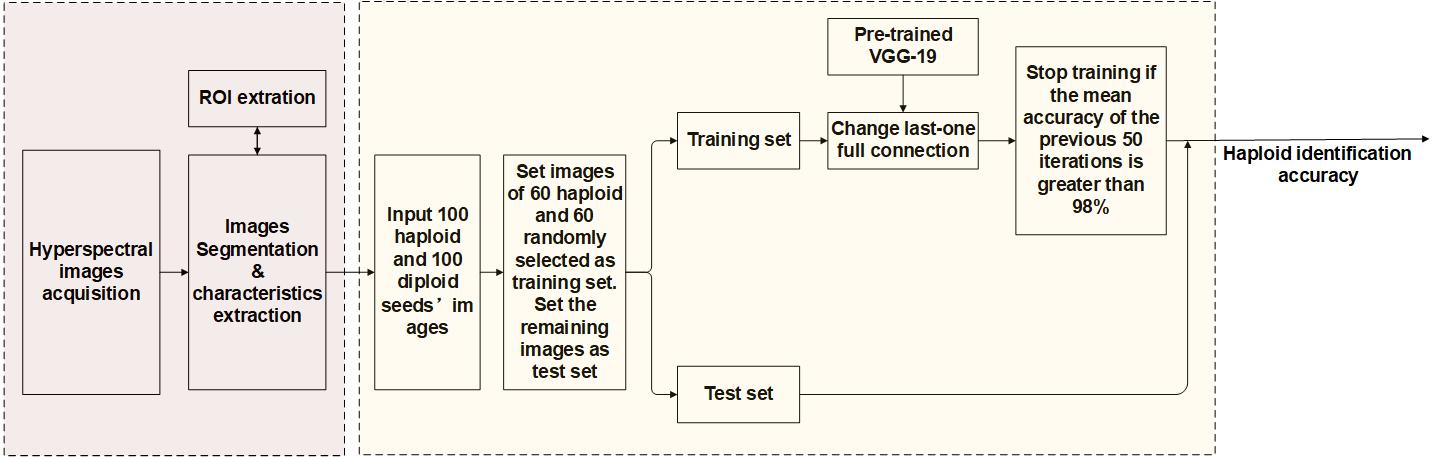}
	\caption{Main steps of identification of haploid maize seeds based improved pre-trained neutral network.}
	\label{fig_sim}
\end{figure*}

\subsection{Transfer Learning}
First, the network goes through sufficient pre-training on Image-Net data set, and the last full connection layer of network is 1000-Dimensional tensor. Although the pre-trained network isn't capable of identifying the maize haploid, it provides excellent initial value to haploid identification network. A good initial value is quite critical to the network training. We have modified the last full connection layer of network (Figure 2), and the output is set as 2, corresponding to haploid and diploid. The network is re-trained using the maize images, and the weight of pre-trained network is fine-tuned, after which the network can be applied to the identification of maize haploid. In the training process, stochastic gradient descent (SGD) parameter optimization is adopted, and the global learning rate is 0.0001. The learning rate of last full connection layer is 0.002, and MiniBatchsize is 90, and MaxEpochs is 500. If the average accuracy of the first 50 iterations reaches 98\%, then the training is terminated early. Cross-validation is performed to work out mean, maximum value and standard deviation of correct recognition rates for five times. The training process of the model contains random process, and the highest value refers to the higher results that the model can achieve in multiple training processes. The experimental process is shown in Figure 3.

\begin{figure}[!t]
	\centering
	\includegraphics[width=2.5in]{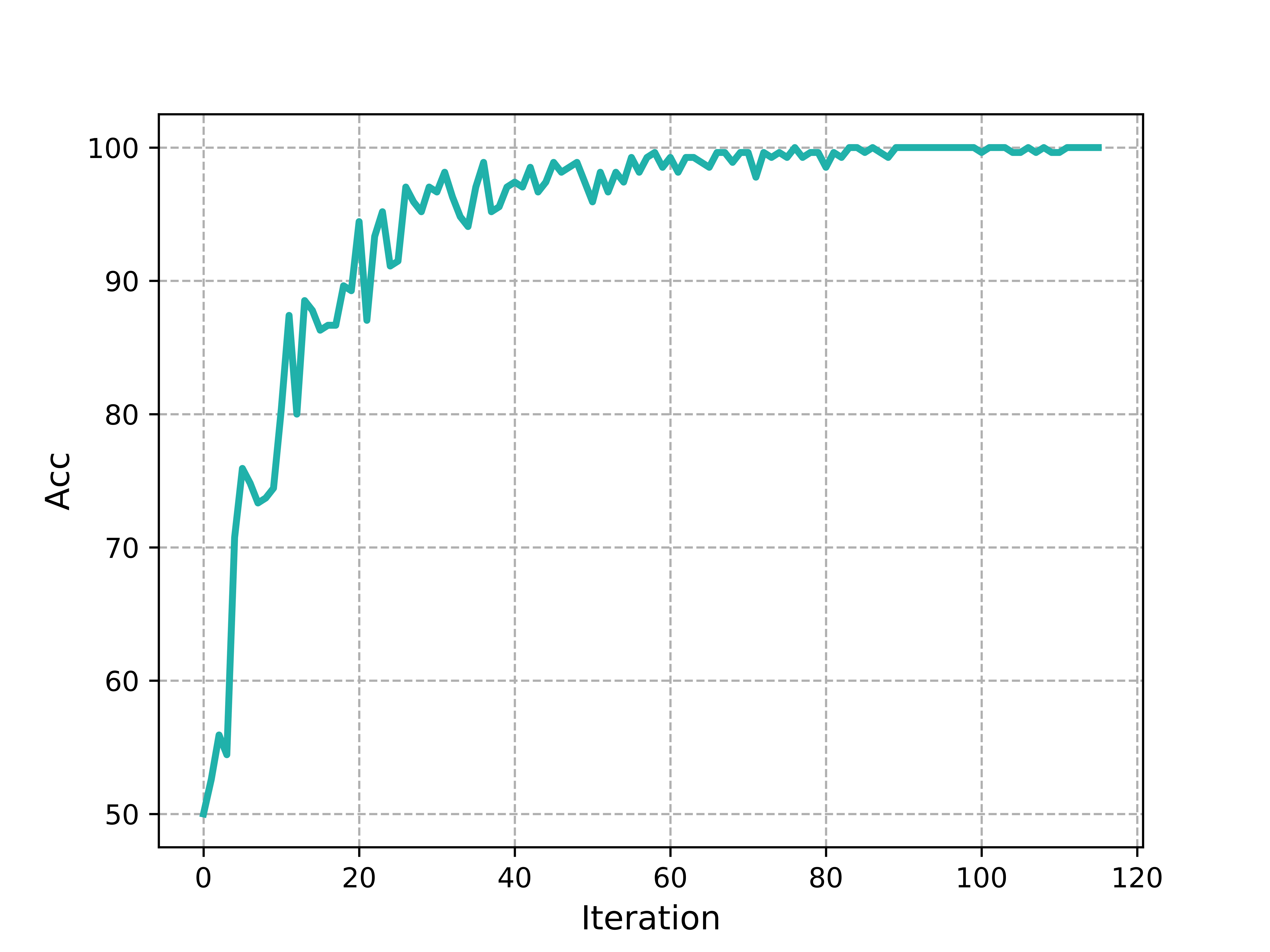}
	\caption{The training accuracy convergence of network on train set, which started converging at 60 times iteration and finally stop at 113 times.}
	\label{fig_sim}
\end{figure}
\begin{figure}[!t]
	\centering
	\includegraphics[width=2.5in]{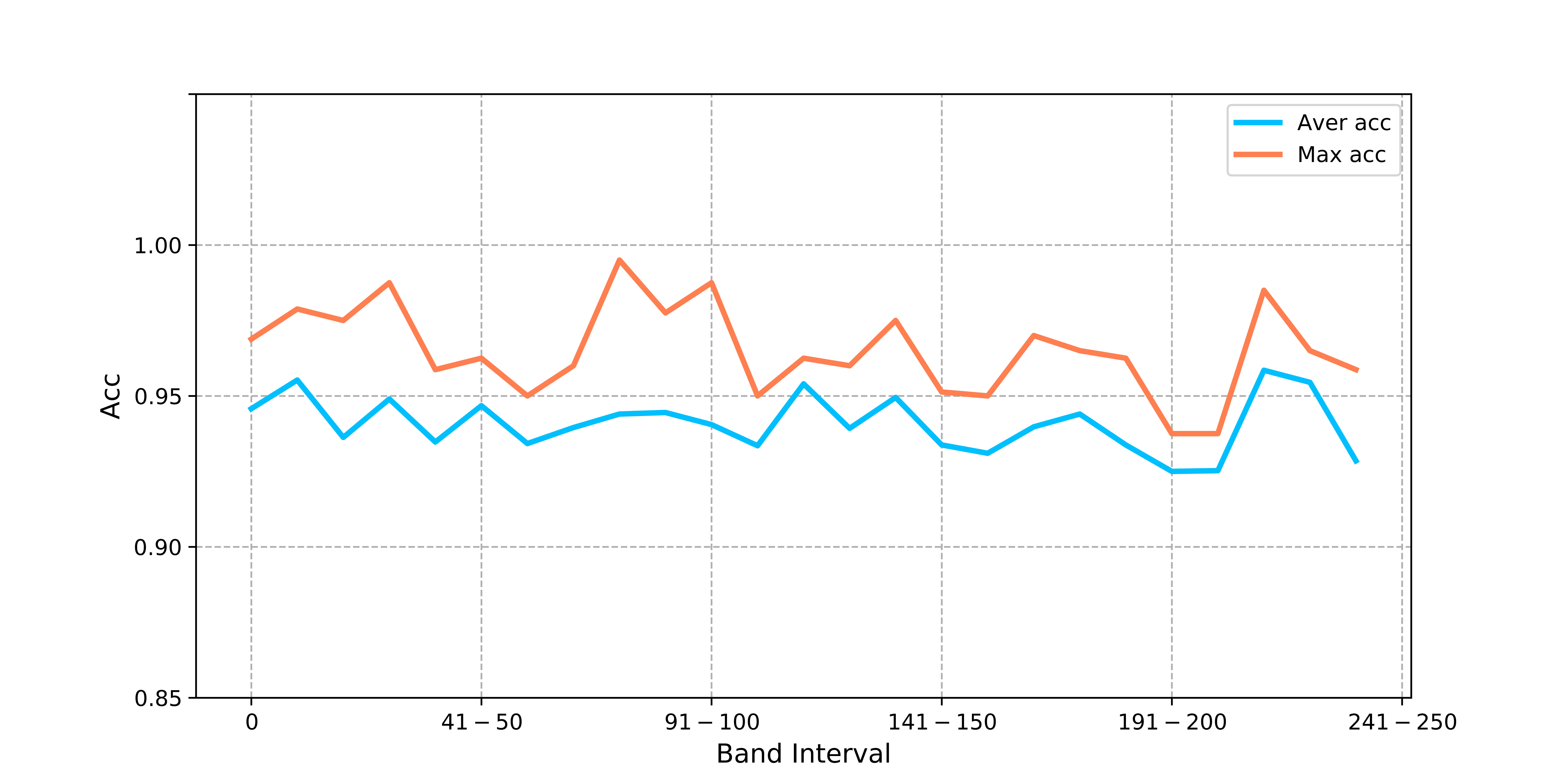}
	\caption{The change of average and maximum accuracies of 25 groups, each of which includes 10 wavebands, during five times cross-validations.}
	\label{fig_sim}
\end{figure}
\begin{figure}[!t]
	\centering
	\includegraphics[width=2.5in]{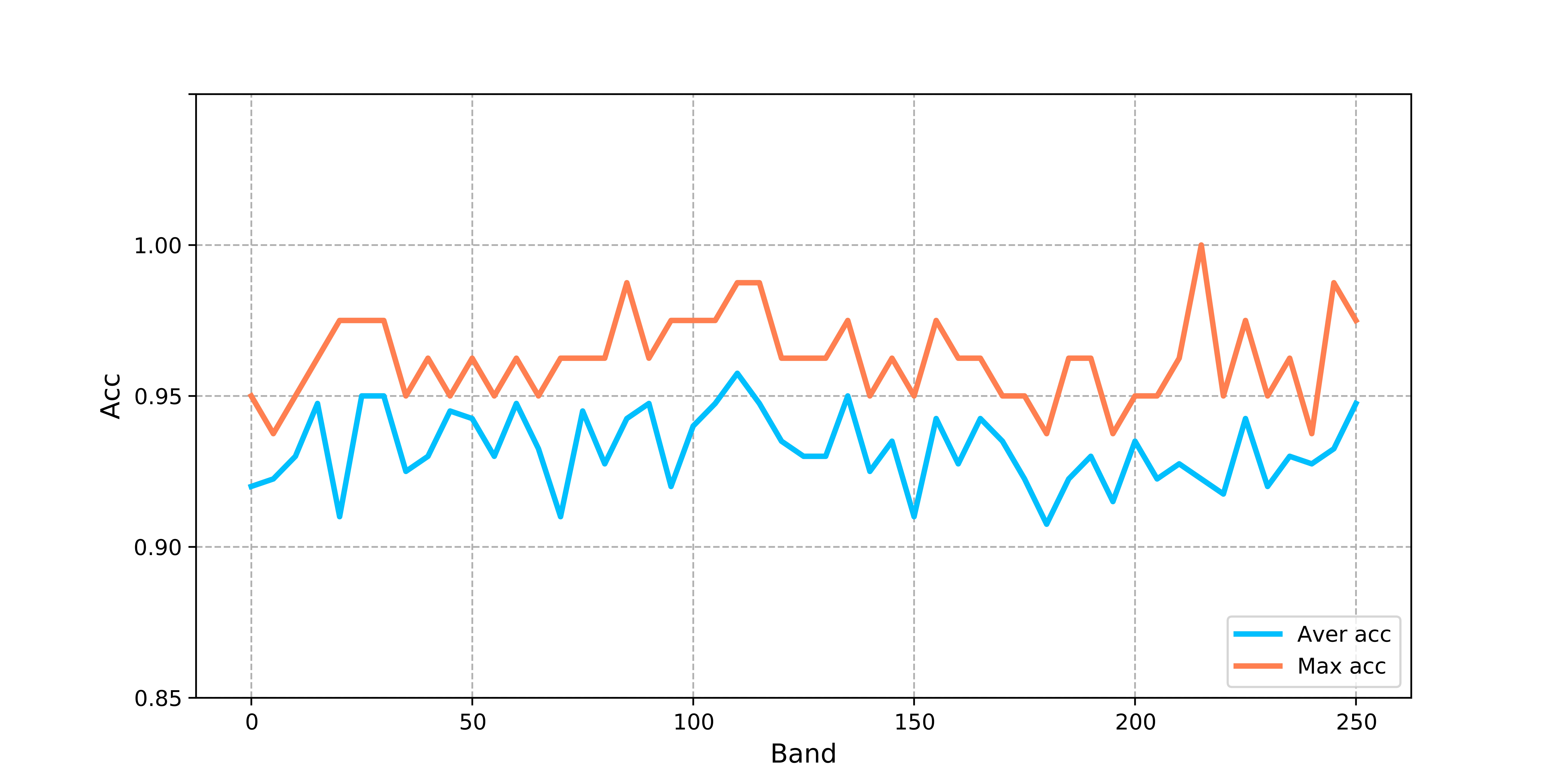}
	\caption{The change of average and maximum accuracies of each single wavebands during five times cross-validations.}
	\label{fig_sim}
\end{figure}
\begin{table}[!t]
	\renewcommand{\arraystretch}{1.3}
	\caption{MODELS FOR IMAGES OF DIFFERENT RANGE OF WAVE LENGTH }
	\label{table_example}
	\centering
	\begin{tabular}{c c c}
		\hline
		\bfseries Range of wave length(nm) & \bfseries Ave acc(\%) & \bfseries Ave consumed time(s) \\
		\hline
		862.9-1704.2 & 96.32 & 338.62\\
		1593.5-1622.1 & 95.85 & 135.43\\
		1249.1 & 95.75 & 65.18\\
		\hline
	\end{tabular}
\end{table}
\section{Results and Discussion}
 The experimental result of modeling of maize hyperspectral images taken at 256bands (862.9-1704.2nm) shows that average accurate identification reaches 96.32\%. The network convergence speed is shown in Figure 4. The network converges after about 60 times of iteration and the training is stopped after 113 times of iteration is finished. The maize haploid can be identified at high accuracy using the model, which meets the actual application requirements. The model training features fast convergence speed, short training time and reduced time cost. Figure 5 shows the experimental result of modeling for 10-wave band images. In the experiment, the hyperspectral images of maize seeds are divided into 25 groups, in which 10 wave bands are classified into one group: 1-10,11-20,21-30, $\ldots$ , 241-250. As seen from the Figure 5, the average accurate identification rate reaches 90\% and above; the highest correct identification rate reaches 93.75\% and above. The average accurate identification rate within the interval of 221-230 (1593.5-1622.1nm) reaches 95.85\%, and the highest correct identification rate within the interval of 81-90 (1135.6-1165.8nm) reaches 99.5\%. The standard deviation of average accuracy among different groups is 0.009, and the standard deviation of maximum value between different groups is 0.015, with relatively small inter-group difference. Figure 6 shows the experimental result of modeling of single-band hyperspectral image of maize seeds In the experiment, one CNN model is built using single-band images every five wave bands, and a total of 51 models are built. The average accurate recognition rate of models reaches 90\% and above, and the highest correct recognition rates exceed 93.75\%, and the average accurate recognition rate of the 115th wave band (1249.1nm) reaches 95.75\%, and the highest correct recognition rate of the 220th wave band (1590.4nm) reaches 100\%. The standard deviation of average accuracy among different groups is 0.012, and the standard deviation of maximum value between different groups is 0.015, with relatively small difference between different wave bands. In three groups of experiments, the average standard deviation shown by cross-validation is 0.018, indicating that the model is stable. Table 2 is the results comparison of above three modeling experiments. As can be seen from the table, the time consumed by single-band image modeling is the shortest. when the data used in modeling is reduced, the correct recognition rate of model doesn't decrease significantly.
 Our research results showed that, after pre-trained VGG-19 network is fine-tuned, the constructed new CNN model can be applied to the identification of hyperspectral image of maize seeds. Among existing researches, through generalization, we can find that, most of models used to identify and sort out seeds are complicated system combination formed by nesting multiple models. For example, Zhang et al.[19] has adopted a kind of PCA-GLCM-LS-SVM iteration combination method when sorting out the seeds. Compared with seeds soring models in the past, the model adopted in the paper doesn¡¯t require other complex algorithms in performing special feature extraction works, as CNN can learn relevant features from the data and classify them. Through the transfer learning, the purpose of pre-trained networks can be expanded, and the problems of sparse sample data can be solved. The generalization performance can be significantly improved with a small number of marked samples. This feature is reflected in the single-band experiment. From the experiment, we can find that, the difference between the highest average recognition rate in the single-band experiment and the average recognition rate in full-wave band experiment is less than 1\%, so the reduction of the training data does not reduce the sorting accuracy of the model.
 Compared with full-wave band model, the single-band model has to face an additional problem, wave band selection. In previous experiments, the important bands are selected according to wave band or Characteristics to improve the accuracy and speed of identification and sorting [20]. The single-band experimental results show that, the standard deviation of average accurate recognition rates among different wave bands is 0.012, and the standard deviation of maximum recognition rates is 0.015. The wave band doesn¡¯t have much effect on the model. Most of the models constructed by the wave band can satisfy the requirement of sorting tasks. No other algorithm is needed in selecting the featured wave band.
 
\section{Conclusions and Future Work}
In the paper, we have built the soring model for identification haploid maize seeds adopting technology of transfer learning and pre-trained VGG-19 network fine-tuned by maize hyperspectral images. The haploid maize seeds can be identified from the maize samples mixed with large amount of diploid seeds with high identification rate. Within the range of 862.9-1704.2nm including 256 bands, hyperspectral imaging system collects feature information of haploid and diploid maize seeds. The model is applicable to both images taken at full-wave bands and images taken at single-band images, the single-band model is featured by low acquisition cost and fast speed. From the perspective of application, it can expand to both accurate maize breeding works and large-scale maize seeds sorting in the seeds breeding.
Our future works include the following (a) Extend the model¡¯s application to other application areas. (b) Test the robustness of the model by applying it in different complicated environment (such as complicated background and illumination condition) (c) Introduce the unsupervised method into the model frame, and improve the autonomous learning ability of features of different categories. Finally, our goal is to integrate our method into the work system that can be used in the site, and to realize automatized operation.

\ifCLASSOPTIONcompsoc
 % \section*{Acknowledgments}
\else
  %\section*{Acknowledgment}
\fi



%\bibliographystyle{IEEEtran}

\end{document}